\documentclass[letterpaper, 10 pt, conference]{ieeeconf}
\IEEEoverridecommandlockouts
\usepackage{array}
\usepackage[T1]{fontenc}
\usepackage{cite}
\usepackage{hyperref}
\usepackage{amsmath,amssymb,amsfonts}
\usepackage{algorithmic}
\usepackage{graphicx}
\usepackage{textcomp}
\usepackage{xcolor}
\usepackage{caption}
\usepackage{algorithm,algorithmic}
\usepackage{lipsum}
\usepackage{gensymb}
\usepackage{mleftright}
\usepackage{bm}
\usepackage{soul}
\usepackage{comment}
\usepackage{mathrsfs}
\usepackage{multirow}
\usepackage{siunitx}
\let\labelindent\relax
\usepackage{enumitem}
\usepackage{microtype}
\usepackage{cleveref}
\usepackage{booktabs}
\usepackage{fvextra} % loads fancyvrb + extras
\usepackage{adjustbox}
\setstcolor{red}
\usepackage{makecell}
\renewcommand{\arraystretch}{1.2}

\usepackage{tcolorbox}
\tcbuselibrary{skins,breakable}
\usepackage{ragged2e}     % for \justifying
\usepackage{microtype}    % nicer justification
\usepackage[T1]{fontenc}
\usepackage{underscore}   % lets you type _ in normal text
\newtcolorbox{promptbox}[1][]{%
  enhanced, breakable,
  colback=black!2, colframe=black!40,
  boxrule=0.5pt, arc=2pt,
  left=8pt, right=8pt, top=6pt, bottom=6pt,
  title=#1,
  before upper=\justifying\setlength{\parindent}{0pt}
}

\usepackage{amsthm}
\usepackage[algo2e,ruled,vlined,linesnumbered,resetcount,norelsize]{algorithm2e}
\makeatletter
\newcommand{\linebreakand}{%
  \end{@IEEEauthorhalign}
  \hfill\mbox{}\par
  \mbox{}\hfill\begin{@IEEEauthorhalign}
}
\makeatother

\newcommand{\wrap}{\mathrm{wrap}_{[-\pi,\pi]}}
\newcommand{\csvhdr}{\texttt{x,y,$\phi$}}

\theoremstyle{definition}

\theoremstyle{plain}

\theoremstyle{remark}

\newcommand{\marginXW}[1]{\marginpar{\color{purple}\tiny\ttfamily#1}}

\newcommand{\AJM}[1]{\textcolor{blue}{#1}}

\def\BibTeX{{\rm B\kern-.05em{\sc i\kern-.025em b}\kern-.08em
    T\kern-.1667em\lower.7ex\hbox{E}\kern-.125emX}}
\begin{document}

%\author{\IEEEauthorblockN}

\title{
Geometry-Aligned LLM Fine-Tuning for Sequential Narrow-Opening Planning
}
\author{
Al Jaber Mahmud and Xuan Wang
\thanks{A. Mahmud and X. Wang are with George Mason University}
\vspace{-0.15em}
} 

\maketitle
\begin{abstract}
We study rigid-body motion planning through multiple sequential narrow openings, which requires long-horizon geometric reasoning because the configuration used to traverse an early opening constrains the set of reachable configurations for subsequent ones. To achieve this, we propose a geometry-aligned large language model (LLM) fine-tuning framework that generates fixed-length, machine-readable waypoint sequences that are both geometrically feasible and coordinated across openings. Our approach uses a bi-level training pipeline. First, we perform failure-driven LoRA supervised fine-tuning (SFT) on human demonstrations, which incorporates structured failure feedback to teach the model common failure modes and enforce the output format. Second, we refine the same LoRA adapters using Group Relative Policy Optimization (GRPO) with geometric verification: each sampled waypoint sequence is densified by a model-based planner and scored with a deterministic geometry-derived reward to achieve continuous-motion feasibility. To validate the effectiveness of our proposed method, we provide both quantitative and qualitative results from simulations. Our method achieves the highest success rate in both in-distribution and out-of-distribution environments and qualitatively exhibits long-horizon geometric reasoning by selecting exit poses that facilitate entry into subsequent openings.
\end{abstract}

\vspace{-0.4em}
\section{Introduction} \label{introduction}

A common problem in manipulation, logistics, and co-transportation~\cite{jang2023motion} is enabling rigid-body navigation through cluttered environments, which often requires passing through multiple narrow openings (e.g., doors or constrained paths between obstacles) in sequence. Unlike single-passage planning, where feasibility can be determined locally, navigating multiple openings requires sequential geometric reasoning: the configuration used to traverse an early opening constrains the set of reachable configurations for subsequent ones.
For example, when carrying a large object through a doorway or co-transporting it in a cluttered space, the object must enter the first opening in a configuration that not only avoids immediate collision but also preserves sufficient clearance and maneuverability to rotate and align for the next passage.
This observation implies that an effective solver must perform \textit{long-horizon geometric reasoning} across sequential openings, selecting early configurations with future constraints explicitly taken into account.

\begin{figure}[t]
    \centering
    \includegraphics[ width=0.35\textwidth]{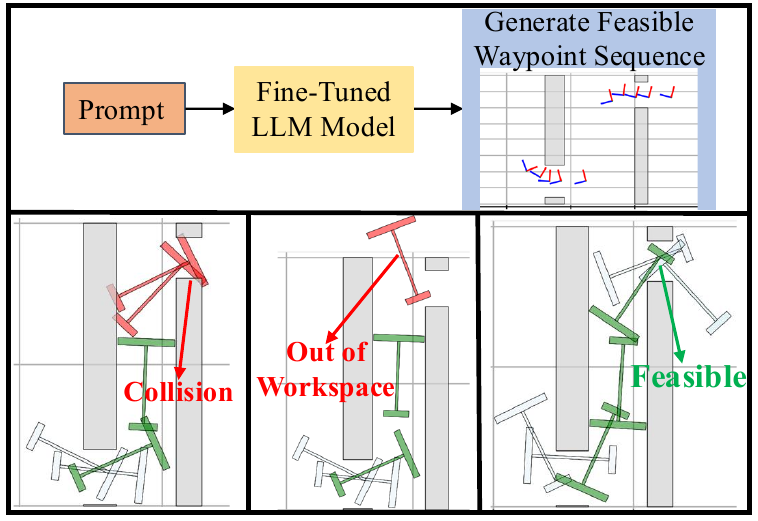}
    \caption{\small 
    Geometry-aligned LLM for Sequential Narrow-Opening Planning and a Motivating Example.
    %\AJM{Overview: Given a prompt containing a symbolic scene description with multiple narrow openings, our fine-tuned LLM outputs a waypoint sequence (top), which is then densified into a continuous trajectory. In narrow inter-opening gaps, the object's poses in the first opening must facilitate entry into the next opening. For example, the I-shaped object must enter the right opening with the shorter bar in front to achieve a feasible trajectory, as shown in (c). Because the inter-opening gap is narrow, the object cannot reorient within it. Thus, when passing through the left opening, the object must enter and exit with a configuration such that the shorter bar is in front.
    %Methods that plan openings independently or optimize only local, short-horizon collision avoidance may overlook this coupling between inter-opening poses and may produce a globally infeasible trajectory.
    %If the I-shaped object enters the left opening with the longer bar in front, it can pass through; however, if it attempts to enter the second opening with the longer bar in front, it may collide with the obstacles. Alternatively, if it attempts to reorient in the gap and enter the second opening with the shorter bar in front, it may go out of the workspace.
    %To reduce these geometric failure modes, our approach integrates long-horizon reasoning with structured geometric feedback and optimization-based evaluation to prepare the object for subsequent openings while traversing the current one (c). This can produce a globally feasible trajectory across sequential openings.
    }
    \label{fig_overview}
    \vspace{-2.2em}
\end{figure} 

\textit{Motivating Example:} As illustrated in Fig.~\ref{fig_overview}, the object pose used to pass through the first opening must also support entry into the second one. Here, even though multiple configurations allow traversal of the first opening (lower left), only a limited subset remains feasible for the second opening (upper right), where the geometric constraints are more restrictive. Since the corridor is too narrow to permit reorientation, the choice made at the first opening largely determines feasibility at the second. Therefore, successful planning requires coordinated reasoning over the full sequence of passages, rather than independent local feasibility checks.

% \marginXW{Why not add a middle point between the two, highlighting the fact that the object cannot pass the corridor. }

%For example, carrying a large object through a doorway or co-transporting in a cluttered space requires a careful navigation strategy to avoid collisions. 
%The problem is difficult because the rigid body must fit and pass through each opening in a pose that still leaves enough space to move and rotate for the next one.
%Moreover, if a planner outputs discrete waypoints, then two waypoints can be collision-free, but the rigid body can collide between these waypoints due to swept motion. Therefore, a solver needs (i) long-horizon coordination across sequential openings and (ii) geometric checks for continuous-motion feasibility.

The requirements illustrated by the above example are not fully satisfied by existing model-based or learning-based approaches. Model-based planners can enforce geometric correctness for collision avoidance, but their performance in multiple sequential narrow openings is often sensitive to implementation choices such as workspace discretization, sampling density, heuristics, and cost tuning~\cite{hsu2003bridge,jaillet2008transition}. 
This makes it harder to find globally feasible solutions within a reasonable time budget. On the other hand, if we decompose the task into intermediate subgoals and plan one opening at a time, it can improve efficiency~\cite{liu2010dynamic, chen2002sandros}. However, it may face issues when composing trajectories: locally feasible exit poses from one opening may not admit a feasible further trajectory in subsequent openings.
Learning-based planners can generate waypoints faster at test time~\cite{song2020mpnet}. However, they usually do not provide geometric guarantees for collision-free continuous motion between waypoints. Their performance may drop when the obstacle layout or rigid body shape differs from the training distribution~\cite{chamzas2019using}.
To address the limitations of existing model-based and learning-based planners in sequential narrow-opening navigation, this paper introduces a geometry-aligned LLM fine-tuning framework for motion planning. The framework integrates long-horizon reasoning with structured geometric feedback and optimization-based evaluation.
An overview of the framework is given in Fig.~\ref{fig_overview}. The main contributions are threefold:
\begin{itemize}
    \item 
    We leverage the sequential reasoning capability of large language models and fine-tune an LLM to generate a fixed-length, machine-readable $\mathrm{SE}(2)$ waypoint sequence structured into per-opening waypoint blocks. This structured output interface enables long-horizon coordination across sequential openings by explicitly selecting exit poses that facilitate feasible entry into subsequent openings.

    \item We introduce a failure-driven training loop in which a deterministic geometric verifier evaluates the densified trajectory (from LLM output) and returns the first failing waypoint index together with the violated constraint type (boundary, collision, swept collision, or step-size). This structured feedback is incorporated into Low Rank Adaptation (LoRA) based supervised fine-tuning (SFT), allowing the model to iteratively reduce recurring geometric errors rather than merely imitating demonstrations.
    
    \item We further refine the same LoRA adapters using Group Relative Policy Optimization (GRPO) with geometric verifier, where each sampled waypoint sequence is densified and evaluated by the geometric verifier to produce a deterministic reward. This reward directly reflects continuous-motion feasibility under geometric constraints, thereby optimizing the model beyond waypoint-level validity and aligning generation with full trajectory correctness.
\end{itemize}
We demonstrate the effectiveness of the proposed approach through experiments and comparison with baseline methods.

\section{Related Work}

\subsection{Model-Based Motion Planning}

% \marginXW{the search?, what search.}
% \marginAM{Addressed.}
In model-based motion planning, the planner searches for collision-free paths in configuration space (C-space), where feasibility can be evaluated with deterministic geometric checks. For example, graph-based planners such as $A^*$ and $D^*$ (or $D^*\text{-Lite}$) perform shortest-path search over discretized grids~\cite{warren1993fast, ferguson2006using}. Sampling-based planners such as Probabilistic Roadmaps (PRM) and Rapidly-Exploring Random Trees (RRT/RRT-Connect) explore by randomly sampling states and connecting nearby samples to approximate connectivity in free space~\cite{
% li2023sample, 
ruan2022efficient, kuffner2000rrt}.
Both graph-based and sampling-based methods face challenges in narrow-opening problems, where feasible motions lie in thin regions of C-space and often require precise orientations to pass through an opening~\cite{ruan2022efficient, salzman2014power}. 
Thus, uniform path search or random sampling may require extensive tuning of key parameters such as grid resolution, sampling budget, connection radius, step size, and the difficulty increases with multiple sequential narrow openings due to geometric constraints and non-convexity. As a result, planners may fail to find a globally feasible path within a reasonable time budget.
One way to overcome this is to decompose the whole task into a sequence of subgoals (each opening as an independent goal) and plan one opening at a time~\cite{liu2010dynamic, chen2002sandros}. While this improves efficiency, the subpaths may fail to compose into a globally feasible trajectory since an exit pose that is feasible for the current opening, but leaves insufficient space to reorient and align with the next opening.

\vspace{-0.15em}
\subsection{Learning-Based Planning}

Learning-based planners use data-driven models to predict trajectories or to guide model-based planners by learning where to search in the C-space.
This can reduce the planning effort and computation at test time. For example, MPNet-style architectures map obstacle information and start-goal poses to continuous paths~\cite{song2020mpnet}.
A network can be trained to propose sampling distributions that improve sampling-based planning in challenging regions~\cite{chamzas2019using}. Other approaches learn cost maps that are then optimized by motion planners such as CHOMP, TrajOpt, or grid-based planners~\cite{wulfmeier2016watch, osa2017guiding}.
These methods may output a feasible set of waypoints based on learned heuristics. However, they often provide limited geometric guarantees for the continuous motion between predicted waypoints. In particular, many models output only a discrete set of waypoints and are trained with losses defined only on these discrete outputs. This does not constrain the intermediate poses between waypoints, so swept collisions can occur even when the individual waypoints are valid.
Learning-based planners can also be combined with hierarchical or decomposed planning by predicting subgoals or waypoints for long-horizon tasks~\cite{pertsch2020long, zhu2021hierarchical}. 
However, in many formulations, the learned model is optimized for single-shot prediction or local objective~\cite{chamzas2019using}, such as imitating demonstrations, motion prediction, or biasing sampling distributions. These objectives do not explicitly enforce sequential coordination across multiple constrained passages and are therefore not directly generalizable for long-horizon geometric reasoning.
Furthermore, as these models rely on learned heuristics, their performance may drop when object geometry or environment layouts differ significantly from the training distribution.

%For example, the model may imitate demonstrations, predict short-horizon motions, or bias sampling in narrow passages.
%These objectives do not explicitly optimize long-horizon planning across sequential openings.
%As a result, the exit pose from one opening may not be compatible with the entry of the next opening, especially when openings are close together, and reorientation space is limited. In addition, learned planners can be sensitive to changes in object geometry or environment structure compared to the training dataset.

\vspace{-0.15em}
\subsection{Large Language Model Based Planners}
Large Language Models (LLMs) are being used as high-level planners combined with model-based or learned controllers.
For example, some methods treat planning as language generation, where an LLM outputs a sequence of symbolic actions or subgoals from textual inputs~\cite{huang2022language}. 
%Other methods connect language to robot skills by scoring candidate skills with value functions and selecting actions that are both consistent with given instructions and executable~\cite{ahn2022can}. 
%In navigation, LLMs are often used for global route reasoning and decision-making from high-level instructions~\cite{shah2023lm}.
Despite strong high-level reasoning, LLM outputs alone do not typically provide geometric guarantees.
Collision checking, motion limits, and continuous feasibility are usually handled by downstream geometric modules such as model-based planners, controllers, or verifiers~\cite{huang2023voxposer}.
Moreover, LLMs are not usually trained to directly output structured, geometrically grounded waypoint sequences that ensure collision-free continuous motion~\cite{christiano2017deep, ibarz2018reward}.
Vision-Language Models (VLMs) or Vision-Language-Action (VLA) may seem to improve geometric correctness because they use visual inputs and can capture richer scene context. 
However, VLM/VLA pipelines still do not provide geometric guarantees by default~\cite{zhang2025safevla}.  
In a VLM/VLA, the model must first infer precise geometry from pixels, and any mistakes in this perception step can directly affect the output generation~\cite{
% fei2024vitron, 
li2023evaluating}. 
They also introduce extra perception components and training costs because of the vision encoder and multimodal pretraining~\cite{li2023blip}.
Therefore, instead of using visual inputs, we can provide exact geometric information in text. This approach can decouple planning from perception. Thus, any differences in performance truly reflect the reasoning behind waypoint generation and feasibility rather than visual estimation.

\subsection{Reinforcement Learning with Large Language Models}
Reinforcement Learning from Human Feedback (RLHF) is a common way to align LLM behavior with human-labeled data, such as demonstrations or rankings of candidate outputs. 
In this pipeline, humans compare multiple model outputs for the same input that are used to train a reward model, which can be costly to collect at scale.
% and these comparisons 
% are used to train a reward model, which can be costly to collect at scale. 
The policy is then fine-tuned with policy-gradient methods, such as Proximal Policy Optimization (PPO), to increase the learned reward~\cite{ziegler2019fine, ouyang2022training}.
To simplify the process, 
Direct Preference Optimization (DPO) converts this human-comparison-based learning into a supervised objective, without the need for online Reinforcement Learning (RL) during fine-tuning~\cite{rafailov2023direct}.
More recently, Group Relative Policy Optimization (GRPO) has been used as an alternative to PPO. 
GRPO computes normalized, relative advantages over a group of sampled outputs for the same input. 
This removes the need to train a separate value network and can improve fine-tuning stability and efficiency~\cite{shao2024deepseekmath, guo2025deepseek}. 
In our setting, this is useful as GRPO can optimize the policy from deterministic geometric verifier scores.
Many prior works use rewards that score the overall textual quality of a response, based on human judgments or learned reward models~\cite{stiennon2020learning, ouyang2022training}.
In contrast, we use a deterministic reward computed by a geometric verifier from output validity and geometric feasibility. 
This makes the optimization directly target feasible waypoint generation rather than subjective response quality.

\vspace{-0.3em}
\section{Problem Statement} \label{problem_statement}
We study rigid-body navigation in a 2-D workspace that contains multiple narrow openings in sequence. The objective is to find a collision-free motion from a given start pose to a goal pose while respecting geometric constraints on the object and the environment. A feasible solution must remain inside workspace boundaries, avoid obstacles, and remain safe during continuous motion between discrete poses (i.e., swept collisions). To evaluate feasibility, we use a deterministic geometry engine that densifies the poses under per-step motion limits and verifies boundary, collision, swept-collision, and step-size constraints.

% \vspace{-0.4em}
% \subsection{Workspace and obstacles}
% \label{subsec:workspace}

\noindent\textbf{Workspace and obstacles:}
Let the workspace be a closed axis-aligned rectangle in the plane $\mathcal{W}=[x_0,x_1]\times[y_0,y_1]\subset\mathbb{R}^2$.
% ,
% \begin{align}
% \mathcal{W}=[x_0,x_1]\times[y_0,y_1]\subset\mathbb{R}^2 .
% \end{align}
Let $\mathcal O=\{\mathcal O_i\}_{i=1}^M$ denote $M$ static obstacles, where each obstacle is a closed axis-aligned rectangle $\mathcal O_i=[a^i_0,a^i_1]\times[b^i_0,b^i_1]\subset\mathbb{R}^2$.
% \begin{align}
% \mathcal O_i=[a^i_0,a^i_1]\times[b^i_0,b^i_1]\subset\mathbb{R}^2 .
% \end{align}
The free space is $\mathcal{F}=\mathcal{W}\setminus \bigcup_{i=1}^M \mathcal O_i$.

% \vspace{-0.4em}
% \subsection{Rigid object and pose}
% \label{subsec:object}

\noindent\textbf{Rigid object and pose:}
The rigid object is specified in its body-frame by an ordered sequence of vertices $\mathcal V_{\mathrm{loc}}=\{v_n\in\mathbb R^2\}_{n=1}^N$, where $v_n=(v_{n,x},v_{n,y})$ is the position of a vertex $n$ and $N$ is the total number of vertices.
The ordering follows the polygon boundary, so consecutive vertices $(v_n,v_{n+1})$ form an edge (with $v_{N+1}=v_1$).
A 2-D pose of the object is $q=(R(\phi), (x, y))\in\mathrm{SE}(2)$, where $\phi\in\mathbb R$ is the heading angle and $(x,y)\in\mathbb R^2$ is the translation in world frame, and
$R(\phi)=\bigl[\begin{smallmatrix}
\cos\phi & -\sin\phi\\
\sin\phi & \cos\phi
\end{smallmatrix}\bigr]\in\mathrm{SO}(2)$
is the rotation matrix.
The corresponding world-frame vertices at waypoint $q$ are
\vspace{-0.25em}
\begin{align}\label{eq_defp}
\mathcal V(q)=\left\{\,p_{n} = R(\phi)\,v_n+ [x ~~ y]^\top\mid n=1,\dots,N\,\right\},
\end{align}
with $p_{n}=(p_{n,x},p_{n,y})$.
We define the object as axis-aligned bounding box $\mathcal{B}$ at waypoint $q$,
\vspace{-0.25em}
\begin{align*}
\mathcal{B}(q)=
\big[\min_j p_{n,x},~\max_j p_{n,x}\big]\times
\big[\min_j p_{n,y},~\max_j p_{n,y}\big]\subset\mathbb{R}^2 .
\end{align*}

% \subsection{Rigid object and pose}
% \label{subsec:object}
% The rigid object is specified in its body-frame by an ordered sequence of vertices $\mathcal V_{\mathrm{loc}}=\{v_j\in\mathbb R^2\}_{j=1}^N$, where $v_j=(v_{j,x},v_{j,y})$ and $N$ is the total number of vertices.
% The ordering follows the polygon boundary, so consecutive vertices $(v_j,v_{j+1})$ form an edge (with $v_{N+1}=v_1$).
% A 2-D pose is $q=(R(\phi),\mathbf t)\in\mathrm{SE}(2)$, where $\phi\in\mathbb R$ is the heading angle and $\mathbf t=(x,y)\in\mathbb R^2$ is the translation in world frame, and
% \begin{align}
% R(\phi)=
% \begin{bmatrix}
% \cos\phi & -\sin\phi\\
% \sin\phi & \cos\phi
% \end{bmatrix}\in\mathrm{SO}(2)
% \end{align}
% is the rotation matrix.
% The corresponding world-frame vertices at pose $q$ are
% \begin{align}\label{eq_defp}
% \mathcal V(q)=\left\{\,p^{(j)} = R(\phi)\,v_j+\mathbf t \mid j=1,\dots,N\,\right\},
% \end{align}
% with $p^{(j)}=(p^{(j)}_x,p^{(j)}_y)$.
% We define the object as axis-aligned bounding box $\mathcal{B}$ at pose $q$,
% \begin{align}
% \mathcal{B}(q)=
% \big[\min_j p^{(j)}_x,~\max_j p^{(j)}_x\big]\times
% \big[\min_j p^{(j)}_y,~\max_j p^{(j)}_y\big]\subset\mathbb{R}^2 .
% \end{align}
%\AJM{This is an axis-aligned rectangle that encloses all transformed vertices. It is only used for broad-phase rejection tests and workspace admissibility checks, described in the next subsection.}

% \subsection{Geometric feasibility constraints}
% \label{subsec:constraints}

\noindent\textbf{Geometric feasibility constraints:}
A waypoint $q$ is feasible if it satisfies:

\noindent\textbf{(C1) Boundary constraints:}
A waypoint $q$ satisfies workspace admissibility if all transformed vertices lie in the workspace, i.e.,
$\mathcal{V}(q)\subseteq\mathcal{W}$.
We enforce this check for every state in the densified trajectory. For an axis-aligned rectangular workspace and a finite vertex set, $\mathcal{V}(q)\subseteq\mathcal{W}$ is equivalent to $\mathcal{B}(q)\subseteq\mathcal{W}$. For that reason, we implement the vertex-wise check directly to avoid relying on derived bounding-box computations.

\noindent\textbf{(C2) Obstacle avoidance:}
A waypoint $q$ is collision-free if the object does not intersect any obstacles.
For each obstacle $\mathcal{O}_i$ and waypoint $q$, we use two-stage collision test:
(i) broad-phase rejection test using axis-aligned bounding boxes: if $\mathcal{B}(q)\cap \mathcal{O}_i=\varnothing$, then the object does not intersect $\mathcal{O}_i$ and
(ii) narrow-phase vertex-wise intersection test where we check the transformed vertices $\mathcal{V}(q)=\{p_n\}_{n=1}^{N}$. We declare a collision with $\mathcal{O}_i$ if any vertex lies inside the obstacle rectangle.

\noindent\textbf{(C3) Swept safety via discrete interpolation:}
To detect collisions that may occur between consecutive waypoints, we perform a swept collision check via discrete interpolation~\cite{bentley2022interactive, swaminathan2025sampling}. 
For each waypoint index $k\ge 1$, we define $S$ number of intermediate substeps $\{\bar q_{k,s}\}_{s=1}^{S}$ at uniformly spaced interpolation parameter
% \marginAM{changed $t_s$ to $\eta_s$.}
$\eta_s = \frac{s}{S+1}\in(0,1)$.
We compute each intermediate substep as
$\bar q_{k,s}=\mathrm{Interp}(q_{k-1}, q_k \mid \eta_s)$, where 
$\mathrm{Interp}(\cdot)$ uses linear interpolation for translation and wrap-aware linear interpolation for the heading angle.
We require that every intermediate substep $\bar q_{k,s}$ satisfies the boundary and obstacle constraints (C1)–(C2).
We choose a value for $S$ so that each interpolated substep is small relative to the per-step translation and rotation limits~\cite{swaminathan2025sampling}.
This discrete interpolation check detects penetrations along the swept motion that might not be detected if only the individual waypoints were verified.

\noindent\textbf{(C4) Step-size smoothness:}
We limit how much the object can translate or rotate per step.
Let $\Delta^{\max}_{\mathrm{lin}}>0$ and $\Delta^{\max}_{\mathrm{ang}}>0$ be maximum acceptable translational and angular movements in each step $h$ of a continuous trajectory. To achieve the step smoothness and avoid abrupt movement,
% of the object, 
we enforce
\vspace{-0.5em}
\begin{align*}
\|[x_h ~~ y_h]^{\top}-[x_{h-1} ~~ y_{h-1}]^{\top}\|_2
&\le \Delta^{\max}_{\mathrm{lin}}, \quad \forall h\ge 1 \\
\big|\wrap(\phi_h-\phi_{h-1})\big| &\le \Delta^{\max}_{\mathrm{ang}}, \quad \forall h\ge 1
\end{align*}

\vspace{-0.5em}
\noindent
where $\wrap(\cdot)\in[-\pi,\pi]$ is principal-value angle wrap.

% \noindent\textbf{(C4) Step-size smoothness:}
% We limit how much the rigid-body object can translate or rotate per step.
% Let $\Delta^{\max}_{\mathrm{lin}}>0$ and $\Delta^{\max}_{\mathrm{ang}}>0$ be maximum acceptable translational and angular movements per step. To achieve the step smoothness and avoid abrupt movement of the object, we enforce
% \begin{align}
% \norm{\mathbf t_k-\mathbf t_{k-1}}_2 &\le \Delta^{\max}_{\mathrm{lin}}, \quad \forall k\ge 1 \\
% \big|\wrap(\phi_k-\phi_{k-1})\big| &\le \Delta^{\max}_{\mathrm{ang}}, \quad \forall k\ge 1
% \end{align}
% where $\wrap(\cdot)\in[-\pi,\pi]$ denotes principal-value angle wrap.

\section{Main Approach}
\begin{figure}[t]
    \centering
    \includegraphics[width=0.46\textwidth]{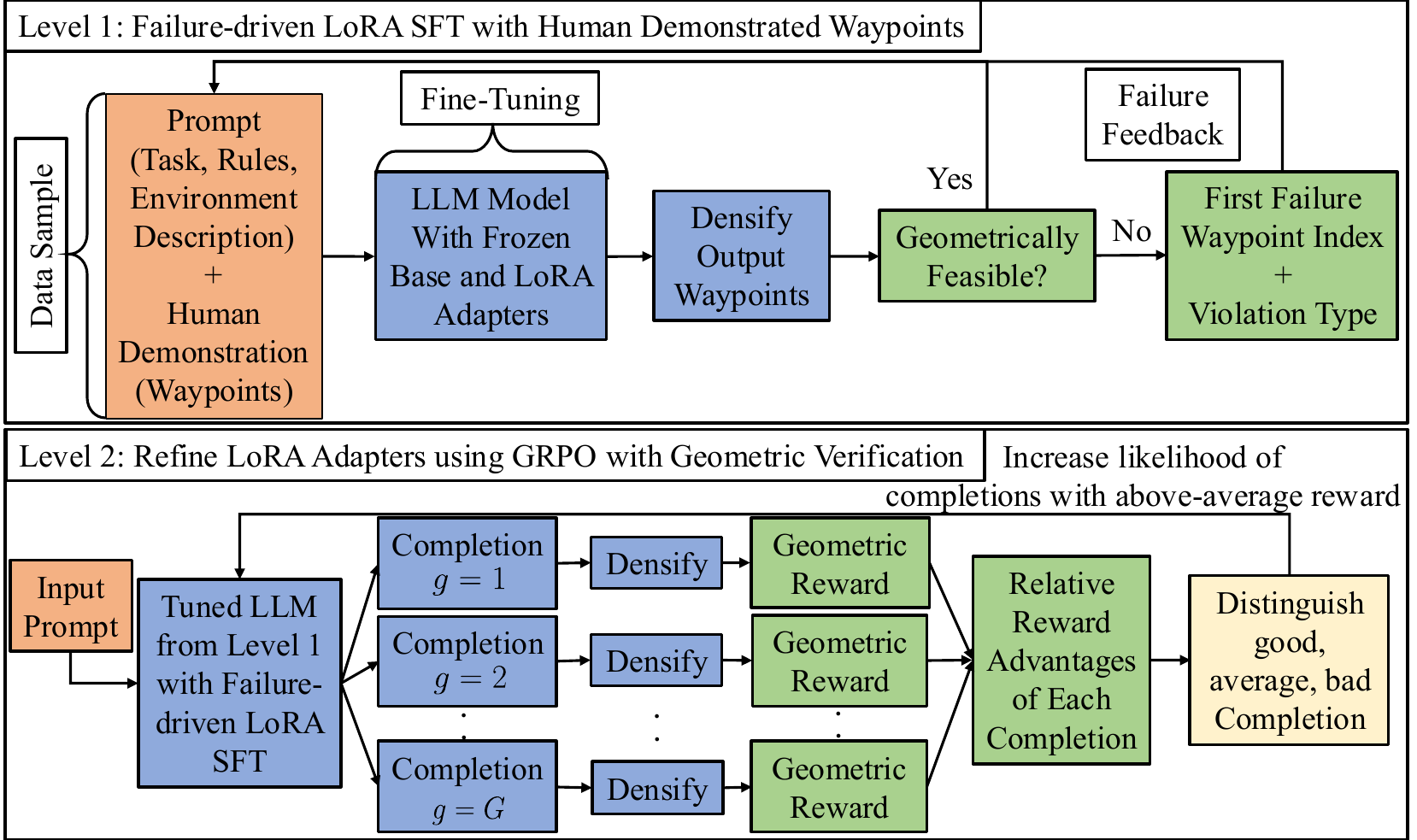}
    \caption{\small Bi-level fine-tuning stage. Level-1: Failure-driven LoRA SFT with human-demonstrated waypoints. Level-2: Refinement of LoRA adapters using GRPO with geometric verification.}
    % \caption{\textcolor{red}{Need to re-organize this figure} Bi-level training framework for geometry-aligned waypoint generation. Level-1 trains LoRA adapters by failure-driven SFT with completion-only masking. Level-2 refines the same LoRA adapters with GRPO using deterministic geometry-derived rewards and relative reward advantages.}
    \vspace{-2.15em}
    \label{demo_overview}
\end{figure}

We fine-tune an LLM to generate fixed-length, 
strictly 
% machine-readable
parseable\footnote{In our context, parseable refers to machine-readable.} 
$\mathrm{SE}(2)$ waypoints from symbolic 2D scenes.
To obtain feasible and strategic waypoint sequences in openings, we collect human demonstrations of rigid-body navigation in synthetic environments (Sec.~\ref{subsec:data_collection}). Each demonstration, together with its corresponding symbolic scene description, is converted into a structured prompt-demonstration pair for fine-tuning (Sec.~\ref{subsec:prompt}).
To fine-tune the LLM, we need two training levels. 
Fig.~\ref{demo_overview} illustrates our bi-level training stages.
In Level-1, we perform failure-driven LoRA
% -based 
SFT (Sec.~\ref{subsec:sft_theory}). Though this teaches the model to output a fixed-length waypoint sequence
% per opening 
and common failure modes, it does not optimize the model for geometric feasibility constraints (C1)-(C4). To address this issue, in Level-2 we perform GRPO with geometric verification, using a geometry-derived scalar reward to refine the LoRA adapters (Sec.~\ref{subsec:grpo}).

%\st{Fig.
% ~\ref{demo_overview} 
%summarizes our bi-level framework and training stages. We first collect human demonstrations of rigid-body navigation in synthetic environments using a Python interface 
% (Sec.~\ref{subsec:data_collection})
%. Each demonstration is converted into a strict prompt-to-CSV training pair 
% (Sec.~\ref{subsec:prompt})
%. This step ensures reliable parsing and enables deterministic geometric verification. In the first training stage (Level-1), we perform LoRA-based SFT
%with completion-only masking, augmented with structured first-failure feedback (failure index and violation type) 
% (Sec.~\ref{subsec:sft_theory})
%. In the second training stage (Level-2), we refine the same LoRA adapters with GRPO. 
%Here, each sampled CSV completion is densified and verified, and a deterministic geometry-derived scalar reward is %used to refine only the LoRA adapters using relative advantages within the sampled group 
% (Sec.~\ref{subsec:grpo})
%.}

\vspace{-0.5em}
\subsection{Human Demonstrations Collection}
\label{subsec:data_collection}
%\marginXW{It is not explained why we need data collection. Why we need it? What is it used for?}
%\marginAM{Addressed.}

\vspace{-0.2em}
To fine-tune an LLM to generate feasible waypoints in sequential narrow openings, we require examples that are both (i) feasible under geometric constraints and (ii) strategic in terms of reflecting long-horizon geometric reasoning, such as selecting an exit pose from one opening that is compatible with entering the next.
Because such feasible solutions are difficult to obtain, as discussed in Sec.~\ref{introduction}, in these narrow, sequential settings, we collect human-demonstrated waypoint sequences as supervision for SFT.
%\st{We collect human-demonstrated waypoints using an interactive keyboard interface. The interface records waypoints for a rigid-body object navigating in synthetic 2-D environments with multiple narrow openings. Our goal is to collect a set of semantically meaningful waypoints per opening. Then, a model-based planner can densify these waypoints, and a geometric verifier can check constraints (C1)-(C4). At first,}
%\marginXW{GUI with keyboard input?} 
%\marginAM{Addressed.}
We generate each training scene by defining a bounded 2-D workspace $\mathcal{W}$ and placing axis-aligned rectangular obstacles that form multiple door-like openings. Each opening is created by two rectangles that share the same $x$-interval and leave a narrow free corridor in the $y$-direction.
For each scene, we store the full symbolic description of the environment required by the prompt and verifier. This description includes workspace bounds, obstacle rectangles, object geometry (local-frame vertices), and the start and goal poses.
We collect demonstrations using a Graphical User Interface (GUI). The human controls the object pose $(x,y,\phi)$ using the keyboard. The arrow keys translate the object in the $x$-$y$ plane, and the \texttt{A}/\texttt{D} keys rotate the heading angle $\phi$. To capture strategic intent,  we let the human decide which poses to record. When the human presses \texttt{ENTER}, the current pose is stored as a waypoint and appended to the waypoint list. The GUI also supports clearing the current trial, resetting to the start pose, and saving the recorded waypoints with the header $x, y, \phi$.

%\st{To obtain human demonstrations efficiently, we implement a lightweight Python interface using Matplotlib for 2-D visualization. The user controls the rigid-body object's position and heading angle using the keyboard. The arrow keys translate the object in the $x$-$y$ plane, and the $A$ and $D$ keys rotate the object by changing the heading angle $\phi$. 
%To capture the user's strategic intent at narrow openings, we let the user decide which waypoints to be stored as data.
%When the user presses \texttt{ENTER}, it stores the current pose as a waypoint in our CSV format (\csvhdr{}) and appends it to the waypoint list. The interface also supports clearing the current trial, resetting to the start pose, and saving the recorded waypoint list to a CSV file with the header \csvhdr{}.
%The object is rendered from its local-frame geometry and transformed by planar rotation and translation during interaction. This ensures the collected demonstrations are directly compatible with the prompt format and the geometric verifier used in training and evaluation.}

%\marginXW{Why different fonts?}
%\marginAM{Addressed.}

\vspace{-0.5em}
\subsection{Prompt Construction}
\label{subsec:prompt}

%\st{After data collection, we convert each training sample into a single self-contained prompt. 
%The prompt is designed to make the LLM's output unambiguous. It enforces strict CSV formatting and a fixed number of waypoints per opening. It also includes the geometric context needed to provide high-level guidance in generating waypoints in the openings. 
%We reuse the same prompt structure during both training and evaluation to reduce train-test drift.}

%\marginXW{It is not clear here.  After subsection A, we only have demonstrations, do we call them 'data'? Then what do we mean by training sample? How can we convert training samples into prompt?}
%\marginAM{Addressed.}

%\marginXW{We can provide example prompts in the video?}
%\marginAM{Addressed.}

\vspace{-0.2em}
Each human demonstration collected in Sec.~\ref{subsec:data_collection}, together with the corresponding symbolic scene description (workspace, obstacles, object vertices, start-goal), forms a single training example.
We convert every training example into a structured prompt-demonstration pair $(\Pi_i,\mathcal{Y}_i)$ used for SFT. Here, $\Pi_i$ is a textual prompt, and $\mathcal{Y}_i$ is the human-demonstrated waypoint sequences for that scene $i$.
The prompt first states the task and the required output format. It specifies the waypoint parameterization, coordinate frame, and the fixed-length requirement. This block reduces common chat-style formatting errors, such as missing headers, extra text, or incorrect row counts.
Next, we explicitly include the constraints (C1)-(C4) in the prompt with the numerical per-step limit.
Then, we include a compact, machine-readable scene description with workspace bounds, axis-aligned obstacle rectangles, object vertices in the local frame, and start and goal poses. Although this geometric information is sufficient for deterministic geometric verification, it can be hard for an LLM to interpret narrow passages from only these numbers.
We therefore add a short human-readable summary of the opening structure that includes the coordinate ranges of the opening region and the available gap width between two adjacent openings. This extra context helps the LLM understand the geometry of the obstacles and the available free space.
Finally, we repeat the required output format and number of waypoints per opening. Though this is redundant, it reduces header drift and the incorrect number of rows.

\vspace{-0.5em}
\subsection{Failure-driven LoRA Supervised Fine-Tuning (SFT)}
\label{subsec:sft_theory}

\vspace{-0.2em}
We fine-tune a frozen base LLM, Llama-3.1-8B-Instruct, using LoRA adapters in the attention and MLP projection layers. 
%\st{The base weights remain fixed, and only the low-rank adapter parameters are updated.}
After fine-tuning, we expect the model to output a fixed-length waypoint sequence for each opening from a textual prompt.
This output format enables reliable parsing to build a continuous trajectory and to run geometric verification under constraints (C1)-(C4).
In general, standard SFT performs next-token prediction on the demonstration tokens (including prompt and demonstrated waypoints)
%\footnote{A token is a discrete unit of text or data that an LLM processes. It can represent a full word, part of a word, a character, or a symbol.} 
%\marginXW{Let's explain the token under our context, should it be the waypoints? We can say: 'In our context, a token refers to xxx '}
%\marginAM{It is not only waypoints. It includes the full prompt as well. Mainly the whole data sample with prompt+demonstrated waypoints.}
to learn how to imitate the human-demonstrated waypoints for each given prompt. However, as the supervision is only a token-level imitation of the waypoints, it does not explicitly inform the model where and why its own rollouts violate feasibility constraints. As a result, the model may repeatedly make geometric mistakes, leading to training inefficiency in both accuracy and time.
We therefore add a learning-from-failure
loop using verifier feedback, while keeping the supervision target (human demonstrations) unchanged. 
Specifically, at the end of each epoch, we run the current model on each training prompt, parse its output waypoint sequence, densify it into a continuous trajectory using a model-based planner, and verify feasibility under (C1)-(C4). If verification fails, we record the first failing waypoint index and the violation type (e.g., \texttt{out\_of\_workspace}, \texttt{collision}, \texttt{swept\_collision}, \texttt{step\_size}). 
With this failure note, we construct a single chat-formatted token sequence
$s_i=\texttt{ChatTemplate}(\Pi_i,\ \texttt{note}_i,\ \mathcal{Y}_i)$,
and apply completion-only masking starting at the header $x,y,\phi$ so that gradients are computed only on demonstrations.
Let $m_{i,t}\in\{0,1\}$ indicate whether token $t$ is inside the human demonstrated waypoints region. We set $m_{i,t}\!=\!1$ for tokens in this region and $m_{i,t}\!=\!0$ otherwise. The training objective is
\vspace{-0.8em}
\begin{align}
\mathcal{L}_{\mathrm{SFT}}
~=~
-\sum_{i=1}^{\mathcal{D}}\sum_{t=1}^{|s_i|}
m_{i,t}\,
\log p_{\theta_{\mathrm{SFT}}}\!\left(s_{i,t}\mid s_{i,<t}\right),
\label{eq:fd_sft}
\end{align}

\vspace{-0.5em}
\noindent
where $p_{\theta_{\mathrm{SFT}}}(\cdot)$ is the model distribution with trainable LoRA parameters $\theta_{\mathrm{SFT}}$ (base model frozen), $\mathcal{D}$ is total number of training data, and $\{s_{i,t}\}_{t=1}^{|s_i|}$ are tokens of full chat sequence. 
This objective keeps the target demonstration $\mathcal{Y}_i$ fixed and conditions generation on $\texttt{note}_i$. With this failure-driven SFT approach, the model generates parseable output that maintains the required format and learns from common failure modes.

\subsection{Group Relative Policy Optimization (GRPO) with Geometric Verification} \label{subsec:grpo}

%\marginXW{Important! We said our method is guided by an `optimal control' like evaluation. Where is it? It should be explicitly defined. }
%\marginAM{Addressed.}

As described in Section~\ref{subsec:sft_theory}, failure-driven LoRA SFT improves output format compliance and identifies common failure modes. However, it does not directly optimize the downstream geometric feasibility criteria (C1)-(C4).
%\marginXW{Sentences similar (concise version) to above are missing at the start of the section.} 
%\marginAM{Addressed.}
We therefore further fine-tune the SFT-initialized model using Group Relative Policy Optimization (GRPO), 
where each sampled waypoint sequence is densified into a continuous motion and scored by our geometric verifier to produce a deterministic reward. This reward reflects the feasibility of continuous motion under geometric constraints and aligns generation with full trajectory correctness.
%\st{The GRPO reward is computed deterministically by our model-based local planner and geometric verifier.}
Let $p_{\theta_{\mathrm{SFT}}}$ denote the SFT-initialized policy. During GRPO, we keep the base model frozen and update only the LoRA parameters. We denote the updated (current) model as $p_\theta$, where $\theta$ contains only LoRA parameters. At the beginning of the GRPO stage, $p_\theta=p_{\theta_{\mathrm{SFT}}}$. We first generate prompts containing different environmental setups and denote this new set of prompts as $\{\Pi_j\}_{j=1}^{\mathcal{N}}$ where $\mathcal{N}$ is the total number of prompts. For each prompt $\Pi_j$, GRPO samples a group of $G$ independent completions $\{\mathcal{Y}_{j,g}\}_{g=1}^{G}$ from $p_\theta(\cdot\mid\Pi_j)$. Each completion is expected to be a fixed-length waypoint sequence in $\mathrm{SE}(2)$. 
%\st{If parsing fails due to a missing header, non-numeric tokens, or an incorrect row count, we assign a large negative reward.} 
%\marginXW{I don't quite understand. The format enforcement, isn't that handled by the previous subsection?}
%\marginAM{Addressed.}
%\st{If parsing succeeds, we densify the waypoint sequence using a model-based local planner. This planner enforces per-step motion limits and produces intermediate states for swept checks. We then evaluate the dense trajectory using the verifier under (C1)-(C4). We define the scalar reward as a weighted sum:} \begin{align} \label{eq_grpo_reward} \mathcal{R}_{j,g} ~=~ w_f\,\mathcal{R}^{\mathrm{fmt}}_{j,g} ~+~ w_o\,\mathcal{R}^{\mathrm{open}}_{j,g} ~+~ w_g\,\mathcal{R}^{\mathrm{geom}}_{j,g}, \end{align} \st{where $\mathcal{R}^{\mathrm{fmt}}$ scores strict CSV compliance (exact header \csvhdr{} and exact row count), $\mathcal{R}^{\mathrm{open}}$ scores per-opening waypoint blocks based on whether they lie in the intended opening corridor, and $\mathcal{R}^{\mathrm{geom}}$ scores the feasibility of the densified trajectory under constraints. $w_f,w_o,w_g >0$ are the corresponding weights. With GRPO, we can avoid the need for an explicit value function by normalizing rewards within each sampled group.} 
%\marginXW{I'll help to reduce the length of the following.}
%\marginAM{Thanks. There are a few notations that have already been used to define something else, which I will change once we finalize this subsection.}
We then use a model-based path planner to connect these waypoints, from start to goal pose, to obtain a continuous trajectory. We denote each state in the continuous trajectory as $\tilde q_h = (x_h, y_h, \phi_h)$ and the complete continuous trajectory as $\tau_{j,g}=\{\tilde q_h\}_{h=0}^H$ for the $j$-th prompt and $g$-th completion, where $H$ is total number of states in continuous trajectory. 
We introduce a quantified measure to evaluate the geometric feasibility of the trajectory
under
% corresponding to 
constraints, as follows

\vspace{-0.5em}
\noindent
{\small
\begin{align*}
\mathcal{C}(\tau_{j,g})
= w_{b}\sum_{h=0}^{H} C_b(h)
+
w_{o}\sum_{h=0}^{H} C_o(h)
+
w_{s}\sum_{h=1}^{H} C_s((h-1), h),
% \label{eq:Vtotal}
\end{align*}}

\vspace{-0.5em}
\noindent
where $C_b(\cdot)$, $C_o(\cdot)$, and $C_s(\cdot)$ quantify boundary (C1), obstacle (C2), and step-size (C4) violations of the trajectory with $w_{b},w_{o},w_{s}>0$ being the weights. Since we score dense trajectory states $\{\tilde q_h\}$ directly, we do not need to evaluate the swept safety constraint (C3) explicitly, as it is evaluated when we perform the workspace violation (C1) and obstacle violation (C2) checks. 
To consistently generate comparison groups for GRPO, we further map
the non-negative violation cost $\mathcal{C}(\tau_{j,g})$ into a normalized score in $[0,1]$,

\vspace{-1.45em}
\begin{align}
\mathcal{R}_{j,g}=\frac{1}{1+\alpha~ \mathcal{C}(\tau_{j,g})}, \qquad \alpha > 0.
\label{eq:Qgeom}
\end{align}

\vspace{-0.45em}
\noindent
where $\alpha$ controls the sensitivity of how fast the function decays.
With this formulation, $\mathcal{R}_{j,g}\!\!=\!\!1$ when $ \mathcal{C}(\tau_{j,g})\!\!=\!\!0$ and $\mathcal{R}_{j,g}\!\!\rightarrow \!\!0$ as $ \mathcal{C}(\tau_{j,g})$ increases. This reward helps to provide a quality measure for each generated trajectory from the same prompt. Thus, GRPO can distinguish \emph{good} (high reward), \emph{average} (intermediate reward), and \emph{bad} (low reward) completions without any learned reward model. 
In the following, we explain in detail how the violation terms are defined. 

% \marginAM{Instead of using superscript, changed it to a subscript. From $p^{(h)}_{n,x}$ to $p_{n,h,x}$.}

\noindent$\bullet$~For the workspace violation (\textbf{C1}), define
\vspace{-0.8em}
\begin{align}
C_b(h)=\sum_{n=1}^{N}\big(\delta_{\mathcal{W}}(p_{n,h})\big)^2.
\label{eq:Vbound}
\end{align}

\vspace{-1.3em}
\noindent
where 
\vspace{-0.6em}
\begin{align*} \delta_{\mathcal{W}}(p_{n,h}) = 
& \max(0,x_0-p_{n,h,x})+\max(0,p_{n,h,x}-x_1) \nonumber \\ &+\max(0,y_0-p_{n,h,y})+\max(0,p_{n,h,y}-y_1),
\end{align*}
measuring how far a vertex lies outside the workspace. Here, $p_{n,h,x}$ and $p_{n,h,y}$ are defined the same way as the vertices in~\eqref{eq_defp} corresponding to $x_h, y_h, \phi_h$
for the rigid object pose. 
If  $p_{n,h}\in\mathcal{W}$, then $\delta_{\mathcal{W}}(p_{n,h})=0$
and increases with the level of boundary violation.

\noindent$\bullet$~For the obstacle violation (\textbf{C2}), define
\vspace{-0.6em}
\begin{align}
C_{o}(h)=
\sum_{i=1}^{M}\sum_{n=1}^{N}
\Big(\max\big(0,- d_{\mathrm{s}}(p_{n,h},\mathcal{O}_i, h)\big)\Big)^2.
\label{eq:Vobs}
\end{align}

\vspace{-0.5em}
\noindent
where $d_{\mathrm{s}}(\cdot)$ is a piece-wise signed vertex-to-rectangle distance to quantify the collision level. 
For each obstacle $\mathcal{O}_i=[a_0^i,a_1^i]\times[b_0^i,b_1^i]$, depending on the vertex $p_{n,h}$ is inside or outside of the obstacles, define
\vspace{-0.4em}
\begin{align*}
d_{\mathrm{s}}(p_{n,h},\mathcal{O}_i, h)=
\begin{cases}
d_{o}(p_{n,h},\mathcal{O}_i, h)& \text{if}~p_{n,h}\in\mathcal{O}_i\\
-d_{i}(p_{n,h},\mathcal{O}_i, h)& \text{otherwise}.
\end{cases}
% \label{eq:sdist}
\end{align*}

\vspace{-0.4em}
\noindent
Here, when $p_{n,h}$ is outside of the obstacles, we measure the Euclidean distance between the vertex and the obstacle as,
\vspace{-0.4em}
\begin{align*}
d_{0}(p_{n,h},\mathcal{O}_i, h)=\sqrt{(p_{n,h,x}-c_{n,h,x})^2+(p_{n,h,y}-c_{n,h,y})^2},
% \label{d_out}
\end{align*}

\vspace{-0.4em}
\noindent
where, $c_{n,h,x} = \min(\max(p_{n,h,x},a_0^i),a_1^i)$ and $c_{n,h,y} = \min(\max(p_{n,h,y},b_0^i),b_1^i)$. The min-max formulation projects the vertex position onto $a_0^i,a_1^i$ or $b_0^i,b_1^i$ to obtain the closest point on the rectangle for Euclidean distance computation. 
When the vertex is inside the obstacle rectangles, we define the distance to the closest face of the obstacle rectangle as 
\vspace{-0.4em}
\begin{align*}
d_{i}(p_{n,h},\mathcal{O}_i, h)=\min\{&p_{n,h,x}-a_0^i,a_1^i-p_{n,h,x}, \nonumber \\ 
&p_{n,h,y}-b_0^i,b_1^i-p_{n,h,y}\},
% \label{d_in}
\end{align*}

\vspace{-0.4em}

\noindent$\bullet$~For the step size violation (\textbf{C4}). We define
\vspace{-0.8em}
\begin{align}
C_{s}((h-1),h)&=
\big(\max(0,\Delta_{\mathrm{lin}}(h)-\Delta^{\max}_{\mathrm{lin}})\big)^2
\nonumber \\ &+
\big(\max(0,\Delta_{\mathrm{ang}}(h)-\Delta^{\max}_{\mathrm{ang}})\big)^2.
\label{eq:Vstep}
\end{align}

\vspace{-0.4em}
\noindent
where $\Delta^{\max}_{\mathrm{lin}}>0$ and $\Delta^{\max}_{\mathrm{ang}}>0$ are translation and rotation limits, respectively. The linear and angular changes are,   
\begin{align*}
&\Delta_{\mathrm{lin}}(h)=\|[x_h ~~ y_h]^{\top}-[x_{h-1} ~~ y_{h-1}]^{\top}\|_2, \nonumber \\
&
\Delta_{\mathrm{ang}}(h)=\big|\wrap(\phi_h-\phi_{h-1})\big|.
\end{align*}

\noindent
With~\eqref{eq:Qgeom} being well defined, for each prompt $\Pi_j$, we compute the group mean and standard deviation, $\mu_j=\frac{1}{G}\sum_{g=1}^{G}\mathcal{R}_{j,g}$ and $\sigma_j=\sqrt{\frac{1}{G}\sum_{g=1}^{G}(\mathcal{R}_{j,g}-\mu_j)^2}$. We then define the group-relative advantage $A_{j,g}~=~\frac{\mathcal{R}_{j,g}-\mu_j}{\sigma_j + \varepsilon}$, where $\varepsilon>0$ is a small stabilizer. This relative advantage increases the likelihood of completions with above-average reward among the $G$ candidates for the same prompt. The term $\varepsilon$ prevents the numerical instability when the within-group variance is near zero. With the increment of the likelihood, it may cause excessive drift from the SFT initialization $p_{\theta_{\mathrm{SFT}}}$. To prevent 
it, we add a KL regularizer toward $p_{\theta_{\mathrm{SFT}}}$ with a KL penalty weight $\beta_{\mathrm{KL}}$. For each prompt, the GRPO objective is 
\vspace{-0.6em}
{\small \begin{align} \label{eq_grpo_objective} J_j(\theta) &= \frac{1}{G}\sum_{g=1}^{G} A_{j,g}\,\log p_{\theta}(\mathcal{Y}_{j,g}\mid \Pi_j)
\nonumber \\ 
&- \beta_{\mathrm{KL}}\, \mathrm{KL}\!\Big(p_{\theta}(\cdot\mid \Pi_j)\ \Vert\ p_{\theta_{\mathrm{SFT}}}(\cdot\mid \Pi_j)\Big). 
\end{align}}

\vspace{-0.4em}
\noindent
We maximize $\sum_{j=1}^{\mathcal{N}} J_j(\theta)$ while updating only the LoRA parameters. The log-likelihood term $(\log p_{\theta}(\mathcal{Y}_{j,g}\mid \Pi_j))$ is computed over the completion tokens
(starting at $(x,y,\phi)$). 
%\st{It is consistent with our completion-only setup. This objective makes the model generate fixed-length, strictly parseable waypoint sequences across multiple sequential openings, while staying close to the SFT behavior.}
With this approach, we refine the LoRA adapters to optimize the model beyond waypoint-level validity and align the generation with full trajectory geometric correctness.
Algorithm~\ref{alg:grpo_abs_compact} summarizes this GRPO fine-tuning process.

\begin{algorithm}[t]
\caption{GRPO Fine-Tuning with Geometry-Derived Reward (LoRA-only Updates)}
\label{alg:grpo_abs_compact}
\begin{algorithmic}[1]
\REQUIRE Prompts $\{\Pi_j\}_{j=1}^{\mathcal{N}}$; SFT-initialized policy $p_{\theta_{\mathrm{SFT}}}$; trainable LoRA parameters $\theta$ (base frozen); group size $G$; object vertices $\{v_n\}_{n=1}^{N}$; obstacles $\{\mathcal{O}_i\}_{i=1}^{M}$; workspace $\mathcal{W}$; motion limits $\Delta^{\max}_{\mathrm{lin}},\Delta^{\max}_{\mathrm{ang}}$
\ENSURE Updated LoRA parameters $\theta$

\FOR{each GRPO iteration}
    \FOR{each prompt $j = 1,\dots,\mathcal{N}$}
        \STATE Sample $G$ completions $\{\mathcal{Y}_{j,g}\}_{g=1}^{G} \sim p_\theta(\cdot\mid \Pi_j)$.
        \FOR{each completion $g = 1,\dots,G$}
            \STATE Obtain continuous trajectory $\tau_{j,g}$ from model-based path planner.
            %\STATE Compute boundary violations $C_b(\cdot)$  (Eq.~\eqref{eq:Vbound}), obstacle violations $C_o(\cdot)$  (Eq.~\eqref{eq:Vobs}), step-size violations $C_s(\cdot)$  (Eq.~\eqref{eq:Vstep}).
            %\STATE Compute total violation cost $\mathcal{C}(\tau_{j,g})$ (Eq.~\eqref{eq:Vtotal}).
            \STATE Compute scalar reward $\mathcal{R}_{j,g}$ (Eq.~\eqref{eq:Qgeom}).
        \ENDFOR
        \STATE Compute group statistics: $\mu_j \leftarrow \frac{1}{G}\sum_{g=1}^{G}\mathcal{R}_{j,g}$,\;
               $\sigma_j \leftarrow \sqrt{\frac{1}{G}\sum_{g=1}^{G}(\mathcal{R}_{j,g}-\mu_j)^2}$.
        \STATE Compute $\mu_j$, $\sigma_j$, and advantages $A_{j,g}$ for all $g$.
        \STATE Compute GRPO objective $J_j(\theta)$ (Eq.~\eqref{eq_grpo_objective}).
    \ENDFOR
    \STATE Update only LoRA parameters $\theta$ by maximizing $\sum_{j=1}^{\mathcal{N}} J_j(\theta)$.
\ENDFOR
\STATE \textbf{return} $\theta$.
\end{algorithmic}
\end{algorithm}

\section{Experiments and Results}
\label{sec:results}

In our experiments, each scene specifies the workspace bounds, obstacle rectangles, object geometry, and the start and goal poses. 
We use LLM-based methods to output an $\mathrm{SE}(2)$ waypoint sequence in all openings. Then, a model-based path planner\footnote{We use the $A^*$ algorithm in our approach. However, one can use other methods such as $D^*$, RRT, PRM, etc. We omit the details of this model-based planner, as they are not the main contribution of this paper.} densifies them to obtain a complete trajectory from start to goal pose. We then evaluate the trajectory with the geometric verifier under constraints (C1)-(C4). 
We initialize all LLM variants from the Llama-3.1-8B-Instruct model.

Our proposed method is \textbf{Failure-driven LoRA SFT + GRPO} with geometric verification (Algorithm~\ref{alg:grpo_abs_compact}), which we quantitatively compare against three LLM baselines under the same prompt and verifier: (i) \textbf{Llama-3.1-8B-Instruct} without any task fine-tuning, (ii) \textbf{LoRA SFT} trained on human demonstrations, and (iii) \textbf{Failure-driven LoRA SFT}, which augments structured failure notes (first failing index and violation type). We also compare our approach qualitatively against (i) \textbf{sub-goal decomposed Probabilistic Road Map (PRM)} and  (ii) \textbf{Failure-driven LoRA SFT}.

\vspace{-0.35em}
\subsection{Quantitative Evaluation}

\vspace{-0.25em}
To evaluate in-distribution (ID) performance, we report output-level and geometry-level metrics over 500 environments in Table~\ref{tab:main_results}.  Here, $\Uparrow$ means larger is better and $\Downarrow$ means smaller is better.
% \marginXW{What's the point if using them without definition, I have added for you.}
We compute \textbf{Parse} ($\Uparrow$) across all environments and check whether the output follows the required format (header and a fixed-format waypoint sequence). 
For geometric evaluation, we densify the generated waypoints with a model-based path planner and then run the same deterministic verifier for every method. 
We report \textbf{Success} ($\Uparrow$) as the percentage of environments whose densified trajectory satisfies all constraints (C1)-(C4). 
For unsuccessful trials, we record the first violated constraint along the densified trajectory and use that for specific \textbf{C1-C4 fail} ($\Downarrow$).

\noindent\textbf{Effect of LoRA SFT.}
When we apply LoRA SFT, it primarily fixes the output interface. The parse rate increases from $70.4\%$ to $100\%$. 
This removes formatting failures and yields a large gain in geometric feasibility. The success rate increases from $5.2\%$ to $54.6\%$. 
It also reduces waypoint-level violations as the boundary failures drop from $8.0\%$ to $3.2\%$, and waypoint collision failures drop from $19.6\%$ to $14.4\%$. 
However, continuous-motion geometric feasibility is still the main challenge. Even when the waypoints are collision-free, the interpolated motion can collide. 
For this reason, swept collisions (C3) remain the largest failure mode at $22.8\%$.

\noindent\textbf{Effect of failure-driven SFT.}
Next, as we augment SFT with structured failure information (first failing index and violated constraint type), it further improves geometric feasibility. The success rate increases to $71.0\%$. 
With this failure-driven model, failures drop across all categories. 
It reduces swept collisions from $22.8\%$ to $16.0\%$ and step-size violations from $5.0\%$ to $1.8\%$. 
These results suggest that the failure feedback helps the model avoid geometric error patterns that waypoint-level SFT does not capture.

\begin{table} [t]
\centering
\caption{\small \textbf{Quantitative} comparison of Failure-driven LoRA SFT + GRPO with geometric verification against baselines on in-distribution (ID) and out-of-distribution (OOD) environments (all values in \%). 
% Higher is better for Parse and Success. Lower is better for C1-C4 fail.
}
% \vspace{-1em}
\label{tab:main_results}

\setlength{\tabcolsep}{3pt}
\renewcommand{\arraystretch}{1.15}
\resizebox{0.45\textwidth}{!}{
\begin{tabular}{|
>{\arraybackslash}m{1.45cm}|
>{\centering\arraybackslash}m{0.56cm}|
>{\centering\arraybackslash}m{0.8cm}|
>{\centering\arraybackslash}m{0.40cm}|
>{\centering\arraybackslash}m{0.44cm}|
>{\centering\arraybackslash}m{0.44cm}|
>{\centering\arraybackslash}m{0.44cm}|
>{\centering\arraybackslash}m{0.8cm}|
>{\centering\arraybackslash}m{1.2cm}|
}
% \begin{tabular}{|m{1.45cm}|m{0.56cm}|m{0.8cm}|m{0.40cm}|m{0.40cm}|m{0.40cm}|m{0.40cm}|m{0.8cm}|m{1.2cm}|}
\hline
\multirow{2}{*}{\textbf{Method}} 
& \multicolumn{6}{c|}{\textbf{ID}} 
& \multicolumn{1}{c|}{\textbf{OOD}} 
% & \multirow{2}{*}{\textbf{Geometric Reward}} 
& \multirow{2}{*}{\makecell[c]{\textbf{Avg.}\\\textbf{Geom.}\\\textbf{Reward}\\ $\Uparrow$}}
\\
\cline{2-8}
& \textbf{Parse} $\Uparrow$ & \textbf{Success} $\Uparrow$ & \textbf{C1 fail} $\Downarrow$ & \textbf{C2 fail} $\Downarrow$ & \textbf{C3 fail} $\Downarrow$ & \textbf{C4 fail} $\Downarrow$ & \textbf{Success} $\Uparrow$& \\
\hline
Llama-3.1-8B-Instruct & 70.4 & 5.2 & 8.0 & 19.6 & 56.0 & 11.2 & 2.1 & 0.09$\pm$0.16 \\ 
\hline 
+ LoRA SFT & 100 & 54.6 & 3.2 & 14.4 & 22.8 & 5.0 & 10.4 & 0.43$\pm$0.27 \\ 
\hline 
\makecell[l]{+ Failure\\Feedback} & 100 & 71.0 & 1.2 & 10.0 & 16.0 & 1.8 & 31.8 & 0.66$\pm$0.32 \\ 
\hline 
+ GRPO & \textbf{100} & \textbf{92.6} & \textbf{0.0} & \textbf{2.0} & \textbf{5.4} & \textbf{0.0} & \textbf{77.3} & \textbf{0.94$\pm$0.18} \\
\hline
\end{tabular}
}
\vspace{-2.2em}
\end{table}

\noindent\textbf{Effect of GRPO.}
Finally, we apply GRPO on top of failure-driven SFT. 
This gives the best performance as we directly optimize a deterministic verifier-based reward that matches our evaluation criteria. 
The success rate increases from $71.0\%$ to $92.6\%$\, and the constraint violations drop sharply. 
C1 and C4 become $0.0\%$, C2 drops to $2.0\%$, and C3 drops to $5.4\%$. 
Overall, the results show a clear progression in terms of geometric feasibility. SFT enforces the output interface and improves waypoint placement. Failure-driven SFT reduces verifier-detected geometric errors. 
GRPO further reduces the continuous-motion failures and produces reliably feasible densified trajectories with in-distribution environments.

\begin{figure}[t]
    \centering
    \includegraphics[width=0.49\textwidth]{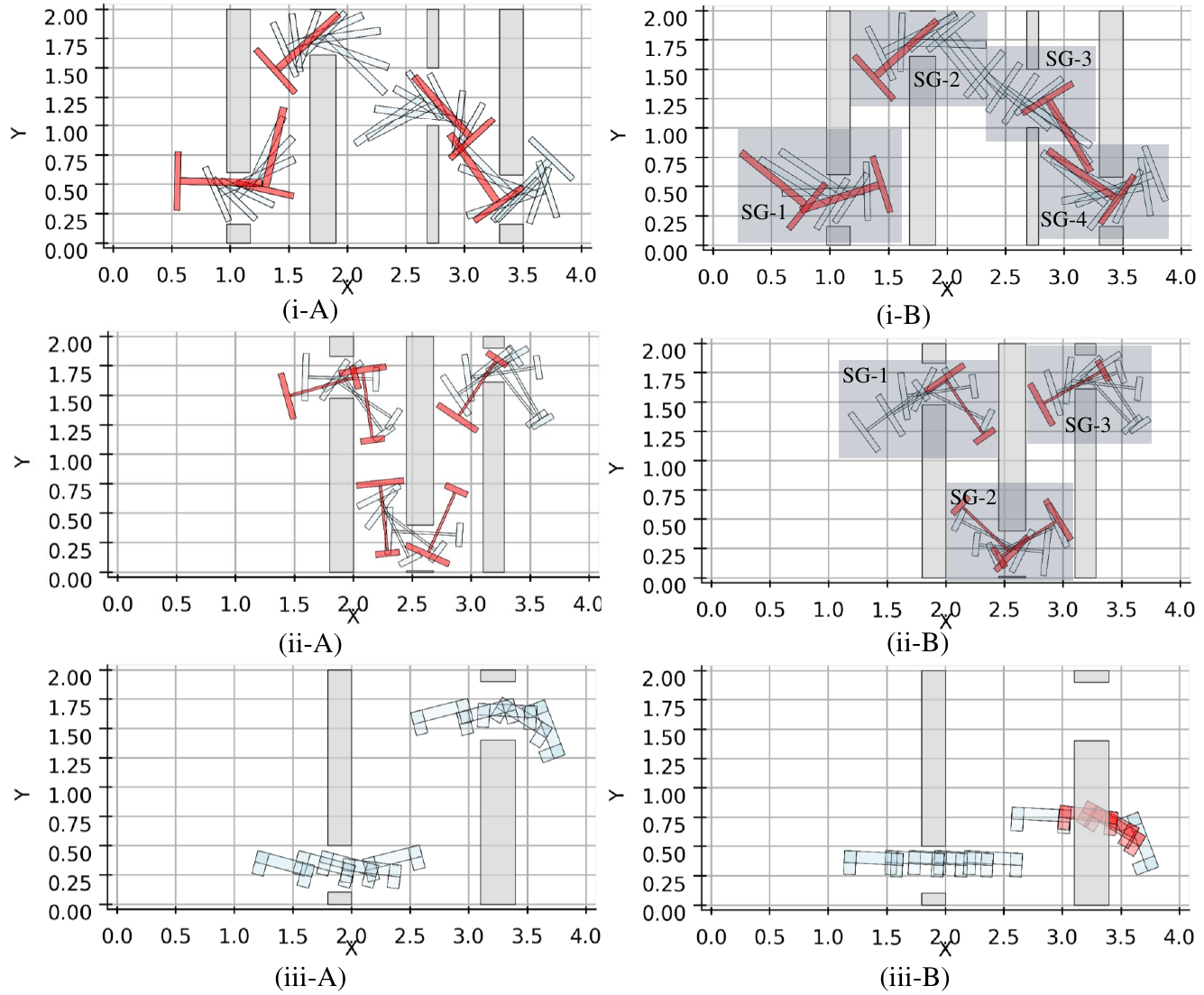}
    \caption{\small \textbf{Qualitative} comparison of methods in terms of Geometric Correctness.
    Gray rectangles are obstacles forming sequential openings.
    \textbf{(i-A and ii-A)} \emph{Failure-driven LoRA SFT + GRPO with geometric verification}: produces exit poses that are already oriented toward the heading needed for the next opening (\textbf{red}).
    \textbf{(i-B and ii-B)} \emph{sub-goal decomposed PRM}: plans each opening independently using subgoal (SG, shaded) regions that yield locally feasible waypoints but leave insufficient space to align for the next opening in low-gap scenes.
    \textbf{(iii-A, iii-B)} \emph{A failure example}: \emph{Failure-driven LoRA SFT} alone can sometimes produce infeasible waypoints \textbf{(iii-B)}, while adding \emph{GRPO} yields feasible waypoints for the same scene \textbf{(iii-A)}.}
    \vspace{-2.0em}
    \label{results_comparison}
\end{figure}

\noindent\textbf{Generalization to unseen environments.}
In addition to in-distribution evaluation, we test generalization on an out-of-distribution (OOD) set of 1000 environments that are disjoint from the training set.
We generate OOD environments by sampling obstacle layouts and opening configurations outside the training ranges (different inter-opening gaps, opening widths, and start-goal poses), while keeping the same object models and evaluation criteria.
We evaluate all methods with the same prompt format, model-based path planner to densify, and deterministic verifier.
We report \textbf{OOD Success} ($\Uparrow$) in the second last column of Table~\ref{tab:main_results}.
In OOD environments, \emph{Failure-driven LoRA SFT + GRPO} achieves $77.3\%$ success.
This shows that GRPO improves geometric feasibility beyond the training distribution as it directly optimizes a verifier-based reward that matches our end-to-end evaluation after densification. It also reflects the increasing long-horizon geometric reasoning capability of this model.
In contrast, \emph{LoRA SFT} alone generalizes poorly. It drops from $54.6\%$ success in-distribution to $10.4\%$ on OOD.
We expect this drop because SFT mainly learns to match the demonstrated waypoint sequences by next-token prediction, and it does not explicitly optimize feasibility after densification.
% When we add structured failure information, 
With \emph{Failure-driven LoRA SFT}, OOD success improves to $31.8\%$, though it drops from $71.0\%$ in-distribution.
This suggests that first-failure indices and violation types help the model avoid repeated geometric error patterns.
However, many OOD scenes are still hard.
In particular, when consecutive openings are separated by short gaps, success depends on
selecting poses at one opening that not only enable feasibility for that opening but also remain feasible for the next, where geometric constraints may be more restrictive.
Consistent with this requirement, we observe that our proposed method, which integrates GRPO with geometric verification, reduces such failures during densification by generating waypoints in which the exit pose from one opening is biased toward the heading required for the subsequent opening. As a result, it improves overall feasibility and achieves the best OOD success rate. This long-horizon geometric reasoning of our proposed method is also visualized in Fig.~\ref{results_comparison}, and discussed in the qualitative evaluation subsection. 
Finally, Llama-3.1-8B-Instruct baseline achieves a very low success rate on both ID and OOD, as it is not fine-tuned for this specific task.

Additionally, we report \textbf{Average Geometric Reward} ($\Uparrow$) with mean$\pm$standard deviation over all 1500 environments (500 ID + 1000 OOD) in Table~\ref{tab:main_results}. 
This is the normalized reward in Eq.~\ref{eq:Qgeom} that is obtained by evaluating the continuous trajectory under geometric constraints.
% after densifying the generated waypoints.
% used to evaluate the quality of a trajectory, as defined in Eq.~\ref{eq:Qgeom}. 
% After densification of each generated waypoint sequence,  
% % For each generated waypoint sequence, we densify it into a continuous trajectory,
% we evaluate the continuous trajectory under geometric constraints, and then compute the normalized score from the resulting violation cost. 
Thus, a higher score indicates a higher-quality (more feasible) densified trajectory.
The Llama-3.1-8B-Instruct baseline achieves a low reward (0.09$\pm$0.16). 
When we apply LoRA SFT, the reward increases to 0.43$\pm$0.27  as it enforces the output format and improves waypoint placement. 
Then, with failure feedback, it further increases to 0.66$\pm$0.32, which indicates fewer and smaller verifier-detected violations after densification. 
Finally, with our proposed method, we achieve the highest reward of 0.94$\pm$0.18 as we directly optimize the verifier-derived geometric score, thereby improving continuous-motion feasibility.

% \vspace{-0.5em}
\subsection{Qualitative Evaluation}

Figure~\ref{results_comparison} shows qualitative comparisons that explain why our proposed method achieves higher geometric correctness after densification under the same verifier.

\noindent\textbf{(i-A), (ii-A) vs. (i-B), (ii-B):} \emph{Long-horizon geometric reasoning capability.}
We show the same qualitative effect for two different object geometries: a T-shaped object in \textbf{(i)} and an I-shaped object in \textbf{(ii)}.
In both scenes, the gap between consecutive openings is short.
Therefore, it is not enough to generate locally feasible waypoints to pass one opening. The poses of one opening must facilitate the entry and feasibility for the next opening.
In \textbf{(i-A)} and \textbf{(ii-A)}, we observe that our method produces exit poses that are already close to the heading needed for the next opening (highlighted in \textbf{red}).
This reduces the rotation required in the inter-opening gap.
It also reduces the risk of swept-collision (C3) because the densified motion requires smaller heading changes within the gap.
In contrast, \textbf{(i-B)} and \textbf{(ii-B)} show a subgoal-decomposed PRM baseline. 
It treats each opening as a subgoal (SG) region and plans for a single opening at a time.
This produces locally feasible waypoints, but it does not explicitly coordinate exit poses across openings.
As a result, the object can exit an opening with a heading that leaves little space to reorient for the next opening, which can lead to infeasibility after densification in low-gap scenes. This comparison reflects the long-horizon geometric capability of the proposed method.

\noindent\textbf{(iii-A) vs. (iii-B):} \emph{Representative failure case without GRPO.}
Here we compare our proposed method against \emph{Failure-driven LoRA SFT} without GRPO on the same scene.
Without GRPO, this model may still place waypoints inside obstacles, which makes the waypoint sequence infeasible (highlighted in \textbf{red}) in \textbf{(iii-B)}.
With GRPO, when we further refine the adapters using the geometric verifier reward, it helps to generate a feasible set of waypoints for the same scene. Thus, the resulting densified motion remains collision-free in \textbf{(iii-A)}.

% \vspace{-0.5em}
\section{Conclusion}
We studied a rigid-body motion planning problem through multiple sequential narrow openings, where feasibility depends on long-horizon geometric reasoning. 
To address this problem, we proposed a geometry-aligned LLM fine-tuning framework that generates fixed-length, machine-readable waypoint sequences that are both coordinated across openings and verifiable under geometric constraints.
We used a bi-level training pipeline: failure-driven LoRA SFT at Level-1, followed by a refinement of the same LoRA adapters at Level-2 using GRPO with a deterministic reward from geometric verification of densified trajectories. We validated the effectiveness of the proposed approach in simulations both quantitatively and qualitatively. 
Future work will extend the framework into 3-D environments with $\mathrm{SE}(3)$ planning and evaluate it in real-world scenarios.

\bibliography{bibliography}
\bibliographystyle{ieeetr}

\end{document}